\documentclass[10pt,twocolumn,letterpaper]{article}
\usepackage{cvpr}
\usepackage{times}
\usepackage{epsfig}
\usepackage{graphicx}
\usepackage{amsmath}
\usepackage{amssymb}
\usepackage{setspace}
\usepackage{multirow}
\usepackage{subcaption}
\usepackage{adjustbox}
\usepackage{floatrow}
\usepackage{pifont}
\newcommand{\cmark}{\ding{51}}%
\newcommand{\xmark}{\ding{55}}%

\usepackage[pagebackref=true,breaklinks=true,letterpaper=true,colorlinks,bookmarks=false]{hyperref}

\cvprfinalcopy 
\def\cvprPaperID{****} 

\ifcvprfinal\fi
\begin{document}

\pagenumbering{gobble}
\title{Quality Guided Sketch-to-Photo Image Synthesis}

\author{Uche Osahor \hspace{.5cm} Hadi Kazemi \hspace{.5cm} Ali Dabouei \hspace{.5cm} Nasser Nasrabadi \\
West Virginia University\\
{\tt\small \{uo0002, hakazemi, ad0046\}@mix.wvu.edu, nasser.nasrabadi@mail.wvu.edu}
\and
}

\maketitle

\begin{abstract}
  Facial sketches drawn by artists are widely used for visual identification applications and mostly by law enforcement agencies, but the quality of these sketches depend on the ability of the artist to clearly replicate all the key facial features that could aid in capturing the true identity of a subject. Recent works have attempted to synthesize these sketches into plausible visual images to improve visual recognition and identification. However, synthesizing photo-realistic images from sketches proves to be an even more challenging task, especially for sensitive  applications such as suspect identification. In this work, we propose a novel approach that adopts a generative adversarial network that synthesizes a single sketch into multiple synthetic images with unique attributes like hair color, sex, etc. We incorporate a hybrid discriminator which performs attribute classification of multiple target attributes, a quality guided encoder that minimizes the perceptual dissimilarity of the latent space embedding of the synthesized and real image at different layers in the network and an identity preserving network that maintains the identity of the synthesised image throughout the training process. Our approach is aimed at improving the visual appeal of the synthesised images while incorporating multiple attribute assignment to the generator without compromising the identity of the synthesised image. We synthesised sketches  using XDOG filter for  the CelebA, WVU Multi-modal and CelebA-HQ datasets and from an auxiliary generator trained on sketches  from CUHK, IIT-D  and  FERET datasets. Our results are impressive compared to current state of the art.
\end{abstract}

\begin{figure}[t]
\begin{center}
  \includegraphics[width=7.8cm, height = 4cm ]{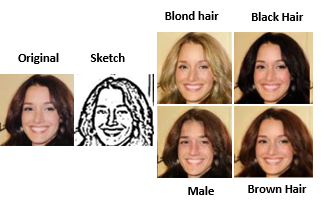}
\end{center}
\vspace{-.70cm}
   \caption{Samples of synthesised images with unique attributes. While training the model, a sketch is conditioned to randomly selected attributes. The final outcome is a group of synthesised RGB images with unique attributes such as male, blond, black and brown hair respectively. }
\label{fig:long}
\label{fig:onecol}
\end{figure}

\begin{figure*}
\begin{center}
  \includegraphics[width=15cm,height=8.0cm ]{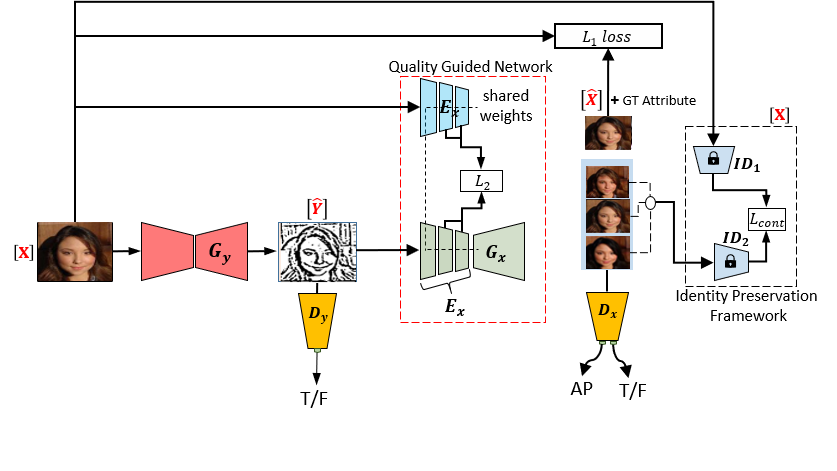}
\end{center}
\vspace{-.8cm}
   \caption{The structure shows the generator and a quality-guided encoder $E_x$ configuration (i.e, green:  $G_y$, $G_x$ and light blue: $E_x$ ) with a 256x256 8-channel input $X$, representing a 3-channel RGB image + 5-channel attribute labels. At  user specified points along the feature extraction pipeline of $E_x$ and $G_x$, the dissimilarity in feature maps are minimized by an $L_2$ loss. The hybrid discriminator (yellow: $D_y$ and $D_x$) extracts features into the last layer where True/False predictions are made and  attribute classification on the 256x256 3-channel is executed to ensure multi-domain adaptation. An identity preserving network comprising of two deep encoders with frozen weights $ID_1$ and $ID_2$ computes a contrastive loss $L_{cont}$ on real and synthesised fake images. }
\label{fig:long}
\label{fig:onecol}
\end{figure*}

\section{Introduction}

Facial sketches drawn by forensic artists aimed at replicating images dictated verbally are a popular practice for law enforcement agencies, these sketches are meant to represent the true features of the individual of interest. Unfortunately, we can’t strictly depend on these sketches as the only means for biometric identification. 

Sketches can be seen as images that contain minimal pixel information bounded by edges that could be translated into photo-realistic images with significant features and pixel content \cite{i2018.00981}. Edge information from such sketches might contain key structural information that aid in providing high quality visual rendition, which is crucial in classifying images as valuable or not. In general, an image could be interpreted as bounded pixels comprising of content and style. Thus, we could either derive sketches manually; from an expert, novice, or through a computer algorithm \cite{5601735}. However, sketches have no standard style template regardless of the mode from which they are drawn or crafted, and as a result, it becomes pretty difficult to translate sketches to images seamlessly \cite{s3332370}. In the sphere of image translation, images can generally be translated from Image to Image (I2I), Sketch to Image (S2I), Edge to Image (E2I), etc. In general, most of the techniques rely on disentangling  the  image into content and style powered by frameworks that adapt a cycled consistency of  conditional generative models \cite{8100115,5995460, you2019pirec,00579}. However, the tendency to obtain rich pixel content with perceptual quality  and discriminative information is still a daunting task, especially for models that move from strictly edge strokes to photo-realistic images.

Scientists and engineers have focused their prowess on developing various solutions that are aimed at translating images  from one domain to another, such solutions could be related to image colorization, style transfer, multi-domain attribute learning and joint optimization \cite{10593-2_13,46487-9_40,2601101,2016.265,vtalreja_globalSip_2018}. These aforementioned techniques utilize models that are either supervised or unsupervised, whose inputs are fed with  either conditioned or unconditioned variables.

Deep Convolutional Neural Networks (DCNN) have evolved into a powerful tool that has made significant advancements in the machine learning and computer vision community and has become valuable in designing machine learning models used to solve image translation problems \cite{vtalreja_ICC,vtalreja_globalSip_2017}.
Their combination with generative models including generative adversarial networks (GANs) \cite{8100115,8347106,inproceedings,8411196}, variational auto-encoders \cite{kingma2013auto}, and auto-regressive models \cite{oord2016pixel} have also achieved significant performance in learning and modeling various data distributions.

Our proposed model synthesizes sketches to produce high-resolution images with a multi-attribute allocating generator that learns to generate images of the same identity but with different target attributes as shown in Figure 1. The final model is a GAN network that transitions from sketches to visually convincing images. To improve quality as compared to state of the art, we introduced a quality-guided network that  minimizes  the perceptual dissimilarity of the latent space embedding of the synthesized and real image at different sections in the network and an identity preserving network that maintains the biometric identity. \cite{Taigman2014DeepFaceCT}.

\begin{itemize}

\item We adopt a single generator for the task of sketch-to-photo synthesis which generates a group of unique attribute guided images of the same subject without compromising the identity of the subject \cite{FANG2020107249}. We also incorporated a verifier that maintains the identity of each subject regardless of the attribute guided image.

\item We develop a single hybrid discriminator that predicts and distinguishes real and synthesized photos according to the set of desired attributes.

\item We adopt a quality-guided encoder which minimises the dissimilarity of the projected latent space of any pair of real and fake images to improve quality.


\end{itemize}

\begin{figure*}
\begin{center}
  \includegraphics[width=17cm ,height =7cm ]{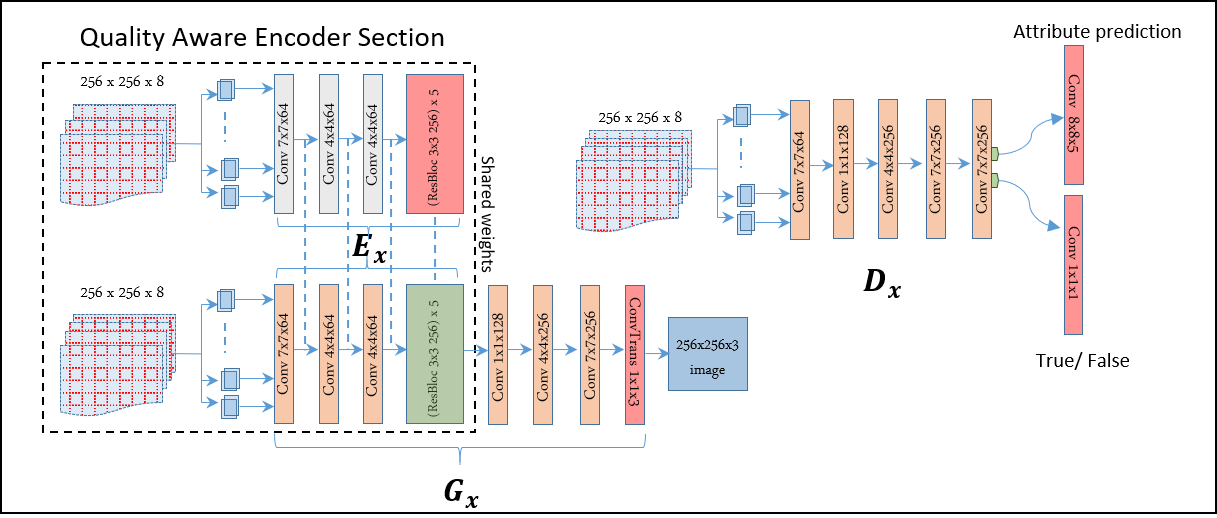}
\end{center}
 \hspace{1.5cm} (a) Generator  \hspace{3.5cm} (b) Discriminator 
   \caption{The proposed architecture is an approach to synthesize realistic images from sketches. (a) The structure shows the generator configuration with a 256x256 8-channel input  representing a 3-channel RGB image + 5-channel attribute labels coupled with the quality guided encoder $E_x$. At  user specified points along the feature extraction pipeline, feature and the corresponding features from $G_x$ interact to minimise an $L_2$ dissimilarity loss. (b) The hybrid discriminator extracts features into the last layer where True/False predictions are made and  attribute classification is executed to ensure multi-domain adaptation.}
\label{fig:long}
\label{fig:onecol}
\end{figure*}

\section{Related Work}
{\bf Generative Adversarial Networks.} Adversarial networks \cite{NIPS2014_5423} have shown great possibilities in the image generation sphere. The main idea behind GANs is to develop an adaptive loss function that improves simultaneously with the generator model. This loss function is formalized by a trainable discriminator which aims to distinguish between the real and generated samples. During the training process, the generator learns to produce more realistic samples in order to fool the discriminator. Recent works showcase the inclusion of conditional constraints \cite{yan2016attribute2Image, vinyals2016matching} while training GANs on images, text, videos and 3D objects and a medley of the aforementioned concepts. Despite the proven capacity of GANs in learning data distributions, they suffer from two major issues, namely the unstable training and mode collapse. 

{\bf Sketch to  image synthesis (S2I).} Sketch to image synthesis can be classified into indirect retrieval and direct synthesis. Indirect image retrieval techniques attempt to solve the sketch to image problem by trying to reduce the domain gap between sketches and photos. However, this approach becomes problematic especially when considering unaligned dataset pairs. Few techniques \cite{Sangkloy_2017_CVPR} have incorporated GAN into S2I synthesis by applying dense sketches and color stroke as inputs. However, complications arise for the GAN architecture due to its inability to accurately assign colors within the boundaries of the object in the image frame. Sketchy-GAN \cite{i2018.00981} proposed a GAN-based end-to-end trainable  S2I approach by augmenting a sketch database of paired photos and their corresponding edge maps. 

{\bf Sketch-based image retrieval.}
Most image retrieval methods \cite{Eitz_2012, Heusel2} use bag of words representations and edge detection to build invariant features across both domains (sketch and RGB images). However this approach failed to form rich pixel image retrieval and also failed to map from badly drawn sketches to photo boundaries. Zhu \textit{et al.} \cite{CycleGAN2017} attempted to solve this problem by treating the image retrieval as a search in the learned feature embedding space. Their approach showed significant progress by registering more realistic fine-grained images and instance-level retrieval.

{\bf Image to image translation (I2I).} Isola \textit{et al.}\cite{8100115} proposed the first framework for I2I based on conditional GANs. Recent works have also attempted to learn image translation with little to no supervision. Some concepts enforce the translation to preserve important properties of the source domain data. The cycle consistency approach is based on the rationale that if an image can be mapped back to its original source, the network is forced to produce more fine grained images with better quality. Huang \textit{et al.} \cite{huang2018multimodal} explored the latent space such that corresponding images in two different domains are mapped to the same latent code; a key struggle for the I2I translation problem is the diversity in translated outputs. Some works tried to solve the problem by simultaneously generating multiple outputs from the conditioned input variables. However, the outputs were limited to a few discrete outputs.

\subsection{Preliminaries}
GANs are generative models that learn the statistical distribution of training data \cite{NIPS2014_5423}, which permits the synthesis of data samples from random noise ${z}$ to an output image $\hat{x}$. Such networks can also be translated in a conditional state that depends on an input image ${x}$.  The generator model $ G(z\sim q(z|x))$ is trained to generate images which are non- distinguishable from their real samples by a discriminator network $D$. Simultaneously, the discriminator is learning to identify the “fake” synthesized images by the generator. The objective function is given as:
\begin{equation}
\begin{aligned}
    \min_{G}\max_{D}V(D,G) &= \mathbb{E}_{x \sim p_{data}(x)}[log D(z|x)] 
     \\  & +   \mathbb{E}_{z \sim p_{data}(z)}[log (1 - D(G(z|x))].
    \end{aligned}
\end{equation}

\section{Multi-Domain Sketch-to-Image Translation}
We established a mapping function between two domains,
$ \{x_i\}_{i=1}^{n}$ $\in \mathcal {X}$ for images and $ \{y_i\}_{i=1}^{n}$ $\in \mathcal {Y} $ for sketches, where $ X=\{x_i\}_{i=1}^{n}$ and $ Y=\{y_i\}_{i=1}^{n}$. Our objective function contains an adversarial loss which moves the distribution $ P(x_i) $ toward the distribution of the target domain 
$ P(y_i) $, an attribute classification loss is integrated into the discriminator to identify attributes of interest and a cyclic constrain that maps a target to their original domain $ G_{x}(x) = P(x_i{\rightarrow \hat{y_j}} \vert x_i) $  and $ G_{y}(y) = P(\hat{y_j}{\rightarrow \hat{x_i}} \vert y_j) $ is implemented, where $\hat{x_i}$ and $\hat{y_i}$ represent reconstructed fake images in each respective domain ($\mathcal{X},\mathcal {Y}$).

To improve the quality of synthesised images, we introduced a quality-guided network  that extracts feature maps from specific layers $\phi_i$ of the encoding section of generator $G_x$ for both real and synthesized pairs and computed an $L_2$ loss to minimize the dissimilarity. We also fused the latent feature maps extracted from the VGG-16 pretrained model for both the synthesised images $ \{ \phi_i(\hat x)$, $\phi_j(\hat y)\}$ and their corresponding original image $ \{ \phi_i(x)$, $\phi_j(y)\}$ to compute the content and style loss between the pair of fake and real images. Identity preservation is achieved for the model with a contrastive loss computation over the real and synthesised image pair using the DeepFace pretrained model, this is to ensure that the synthesised images preserve their verification integrity while training. The resultant architecture is a DCNN structure that aids the network produce quality images \cite{2017629, Hu2020FacialAS}, as shown in Figure 2. With this approach, we eased the burden on the generator and scaled the problem into smaller tasks; making it easier for each intermediate section of the model to identify small challenges like edge detection, color balance, global and local feature learning, yet minimizing computation time. 
\setlength{\textfloatsep}{1.1pt}
\begin{figure}[t]
\begin{center}
  \includegraphics[width=7cm,height =3.4cm ]{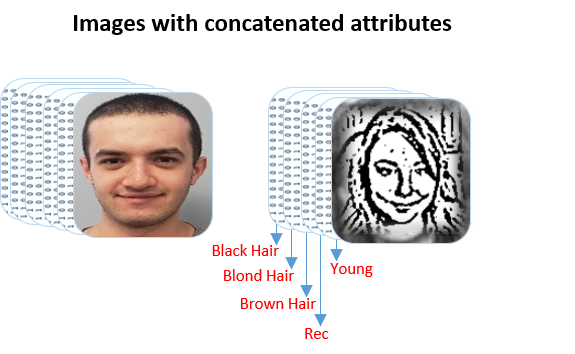}
\end{center}
\vspace{-.5cm}
  \caption{The input template are an attribute concatenation representation for all the training data for both RGB images and sketches. We spatially replicate the labels into image layers and concatenate them with the input images, forming a total of 5 channels for attributes [ black hair, blond hair, brown hair, young and rec (ground-truth label)] + 3 channels for sketch and RGB images, respectively.
  }
\label{fig:long}
\label{fig:onecol}
\end{figure}

Figure 3 illustrates the internal structure of the generator and hybrid-discriminator. 
Adversarial losses \cite{NIPS2014_5423}  were applied to the mapping function $ G_y(x): \mathcal{X} \rightarrow \mathcal{Y}. $ The generator $ G_y $ is trained to synthesize images $ G_y(x)$.

\noindent
The objective function is defined as: 
    \begin{equation}
    \begin{aligned}
         \mathcal {L}_{adv}(G,D,X,Y)  &=  \sum_{i=0}^{m-1}  
         \mathbb {\{E}_{y \sim p_{data}(y)}[log D_i(y)] \\  & +  
         \mathbb{E}_{x \sim p_{data}(x)}[log (1- D_i(G_i(x))]{ } \\ & +
     \sum_{i=0}^{m-1} \mathbb{E}_{x \sim p_{data}(x)}[log D_i(x)] \\ & +  \mathbb{E}_{y \sim p_{data}(y)}[log (1- D_i(G_i(y)))]{\}}.
    \end{aligned}
    \end{equation}

\begin{figure*}
\begin{center}
\includegraphics[width=17.0cm, height = 5cm]{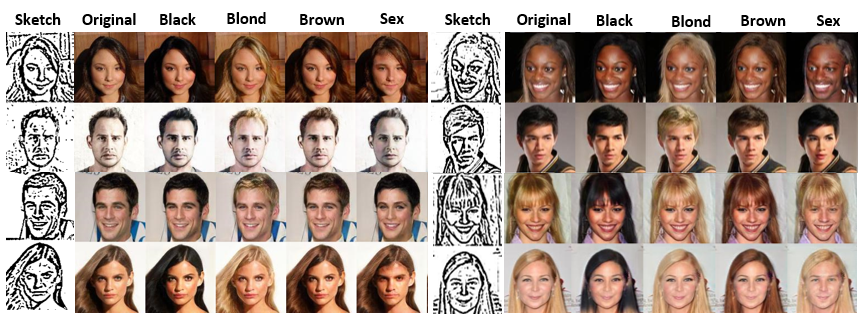}
\end{center}
\vspace{-.5cm}
   \caption{A spectrum of synthesised Images with different target attributes.}
\label{fig:attribute}
\end{figure*}
\subsection{Quality guided network}
Image quality is crucial for sketch to image synthesis, but most GAN networks fail to improve the quality of images due to GANs tendency to only capture a subset of the variation and deep features found in the training data \cite{Karras2017ProgressiveGO}. Hence, applying statistical computations on mini batches has become crucial for improving the overall GAN performance as shown in Figure 5, we use this approach towards improving the quality and variation of the data. The encoder network $E_x$ is initialized to share weights with the  encoding section of the sketch to image generator $G_x$, we use this technique to ensure that the discriminator guides the performance of $E_{x}$ at different resolution stages to improve overall image quality in $G_x$. For each minibatch, we compute the standard deviation for each feature, which we average over all the features and spatial locations \cite{Kingma2017ImprovedVI}, we then project the resolved scalar solution as an extra channel towards the end layers of the discriminator $D_x$. For the intermediate layers of the quality aware network, we fade in new layers smoothly as we transition between layers per minibatch. To ensure the network maintains stability, we locally normalize the feature vector in each pixel to unit length in the generator after each convolution layer. The expression is given as:

\begin{equation}
\begin{aligned}
    b_{xy}
    &= 
    \frac{a_{x,y}}{\sqrt{{\frac{1}{N}} \sum_{j=0}^{N-1} ({a_{x,y}})^2 + \epsilon }  },
 \end{aligned}
\end{equation}
where $\epsilon$ = $10^8$ $N$ is the number of feature maps, $a_{x,y}$ and $b_{x,y}$ are normalized feature vectors in pixel$(x,y)$, respectively. Finally,  an $L_2$ loss is computed for each of the  corresponding layers $E_x$ and the encoding section of $G_x$.

\subsection {Identity preservation}
To design a robust sketch to image synthesis model,  the identity of individual subjects must be consistent with zero tolerance for identity mismatching \cite{DBLP:journals/corr/abs-1812-01288}, which is obviously crucial for applications such as forensic analysis. Due to the degree of sensitivity of face verification, We incorporated  a pretrained DeepFace verification model, sensitive to the identity features of a face trained with millions of faces. We choose the DeepFace model \cite{Taigman2014DeepFaceCT} because it depends on DCNNs to directly optimize the embedding which is preferable to other techniques that use an intermediate bottleneck layer. We applied the contrastive loss \cite{Hadsell2006DimensionalityRB} from the features extracted from the DeepFace model for pairs of real and synthesised images to maintain the hidden relationship between the original image and the synthesised target image in a common latent embedding space. The contrastive loss ensures that the semantic similar pairs of real and synthesised images share a common embedding while dissimilar images (impostor pairs) are pushed apart.

The contrastive loss is expressed as:
\begin{equation}
    \begin{aligned}
    \mathcal{L}_{cont}(\varphi_i(x_i),\varphi_j(y_i),Y) 
    &= 
    (1-Y)\frac{1}{2}{(D_\varphi)}^2
     \\ & +
    (Y)\frac{1}{2}{(max(0, m - {D_\varphi}))}^2, 
    \end{aligned}
\end{equation}
 where $y_i$ and $x_i$ represent the synthesised and real image respectively. The variable $Y$ is a binary label, which is equal to $0$ if  $ y_i$ and $x_i$ belong to the same class, and equal to $1$ if $ y_i$ and $x_i$ belong to a different class. $\varphi(.)$ denotes the encoding functions of the encoder that transforms $ y_i$ and $x_i$ respectively into a common latent embedding subspace. The value $m$ is the contrastive margin and is used to “tighten” the constraint. $D_\varphi$ denotes the Euclidean distance between the outputs of the functions $\varphi_i(x_i), \varphi_j(y_i)$.

\begin{equation}
    \begin{aligned}
   D_\varphi =  {\left\| \varphi_i{(x_i)} - \varphi_j{(y_i)} \right\|_2},
    \end{aligned}
\end{equation}
If $Y = 0$ (genuine pair), the contrastive loss becomes:

\begin{equation}
    \begin{aligned}
    \mathcal{L}_{cont}(\varphi_i(x_i),\varphi_j(y_i),Y) 
    &= \frac{1}{2} \left\| \varphi_i {(x_i)} -
     \varphi_j{(y_i)} \right\|^2_2,
    \end{aligned}
\end{equation}
and if $Y = 1$ impostor pair, the contrastive loss is given as:
\begin{equation}
    \begin{aligned}
    \mathcal{L}_{cont}(\varphi_i(x_i),\varphi_j(y_i),c(i,j)) 
    &= \\\frac{1}{2} max  {\bigg (0,m -}
    & { \left\| \varphi_i {(x_i)} - \varphi_j{(y_i)} \right\|^2_2 \bigg )},
    \end{aligned}
\end{equation}
The total loss is given as:
\begin{equation}
\begin{aligned}
    \mathcal{L}_{cont}(\varphi_i(x_i),\varphi_j(y_i))
    &=  \\
    & \frac{1}{N^2} {\sum_{i=1}^{i} \sum_{j=1}^{j}} \bigg \{ 
    {\left\| \varphi_i{(x_i)} - \varphi_j{(y_i)} \right\|^2_2} \bigg  \}.
 \end{aligned}
\end{equation}

\subsection {Style transfer loss}
The style transfer loss comprises of the  content and style loss, the content loss is derived from the interaction of an image to the layers of a convolutional neural network trained on object recognition such as VGG-16. When these images interact with the layers, they form representations that are sensitive to content but invariant to precise appearance \cite{2016.265}. The content loss function is a good alternative to solely using $L_1$ or $L_2$ losses, as it gives better and sharper high quality reconstruction images \cite{Tenenbaum2000SeparatingSA}.
The content loss is computed for both  real and synthesized images using a pre-trained VGG-16 network. We extract the high-level features of all the layers of VGG-16 weighted at different scales. The $L_1$ distance between these features of real and synthesised images are used to guide the generators $G_x$ and $G_y$. 

\noindent
The content loss is given as:

\begin{equation}
\begin{aligned}
    \mathcal{L}_{cnt}(\phi_i(x_i),\phi_j(y_i))
    &= 
    \frac{1}{C_p H_p W_p} {\sum_{c=1}^{C} \sum_{h=1}^{H} \sum_{w=1}^{W}} \\
    &
    {\left\| \phi_i{(x_i)_{c,w,h}} - \phi_j{(y_i)_{c,w,h}} \right\|_1}, 
 \end{aligned}
\end{equation}

The style content is obtained from a feature space designed to extract texture information from  the layers of a DCNN. The  style contents are basically correlations between the different filter responses comprising of the expectation over the entire spatial space of the feature maps. These are obtained by taking the gram matrix ${G^\phi}(.)$ to be the $C_j $ x $ C_j$ matrix whose elements are given by: 

\begin{equation}
\begin{aligned}
   {G^\phi}_i(x)_{c,c'}
    &= 
    \frac{1}{C_p H_p W_p} { \sum_{h=1}^{H} \sum_{w=1}^{W}} \\
    &
     (\phi_i{(x_i)_{c,w,h}} ) (\phi_i{(x_i)_{c',w,h}}),
 \end{aligned}
\end{equation}
where $\phi(.)$ represents an output of a layer in the VGG-16 layers, 
where $C_p, W_p$ and $H_p$ represent the layer dimensions.

The style reconstruction loss for both images is thus an $L_1$ loss of each computed gram matrix:

\begin{equation}
    \begin{aligned}
   {G^\phi}_i,_j(x,y)_{c,c'} =  \left\| {G^\phi}_i(x)_{c,c'} - {G^\phi}_j(y)_{c,c'}\right\|_1.
    \end{aligned}
\end{equation}

\subsection{Attribute classification loss}
To ensure a robust sketch-to-image translation process, the desired attributes $c'_i$ were concatenated to each image  $x_i$ \cite{8347106}. This combined input is used to train the generator, making it possible to produce images conditioned on any preassigned target domain attributes $c'$. We achieve this by permuting the assigned target label generation process with respect to the original ground attributes $c$ of the image, this is to ensure that the generated labels are unique from the original. The auxiliary attribute classifier within the discriminator is responsible for injecting the attribute property to the sketch for the image generation process. Figure 4 illustrates the visual representation of the attribute integration process. The combined constraints compute  domain classification losses for both real and fake images. We optimize the discriminatory power of our network to optimize $D$ for real images and $G$ for fake images.

\noindent
The combined losses are expressed as:

\begin{equation}
\begin{aligned}
     \mathcal{L}^r_{cls} =  \mathbb{E}_{x,c}[-log D_{cls}(c|x)], 
  \end{aligned}
\end{equation}

\begin{equation}
\begin{aligned}
       \mathcal{L}^f_{cls} =  \mathbb{E}_{x,c' }[-log D_{cls}(c'|G(x,c))]. 
\end{aligned}
\end{equation}

\subsection{Reconstruction loss}
Adversarial networks are capable of learning mappings between the source and the target domain. Hence, It is possible to reproduce inconsistent images that do not share the same property as the desired output. In this regard, utilizing strictly adversarial losses is not a sufficient approach to arrive at a desired output. Hence, it becomes imperative to ensure that the learned mapping of the network be cycle-consistent. Cycle consistency loss postulates that for an arbitrary input image $x$ from its domain  $ \mathcal{X} $ an image  $\hat{x}$ can be reconstructed; $ x \rightarrow G_{y}(x) \rightarrow G_{x}(G_{y}(x)) = \hat{x} $. Likewise, an image $y$ from a different domain  $ \mathcal{Y} $   can satisfy the same principle backwards as $ y \rightarrow G_{x}(y) \rightarrow G_{y}(G_{x}(y)) = \hat{y}$.  These collective approaches adapt the transitivity rule which is developed as cyclic consistency. A reconstruction loss which  hinges on the cyclic consistency \cite{8100115}  and the perceptual appeal \cite{JohnsonAL16}, obtained by extracting VGG-16 layers $\phi_i$, is applied as an ${L}_1 $ loss between  a real image and its corresponding reconstructed version (fake image). Its general equation is given as:
\begin{equation}
\begin{aligned}
 \mathcal{L}_{rec} = \frac{1}{N^2} \sum_{i=1}^{i}  \sum_{j=1}^{j} \bigg \{ \left\| \phi_i{(x)} - \phi_i{(\hat x)} \right\|_1 
 &+ \\
 \left\| \phi_j{(y)} - \phi_j{(\hat y)} \right\|_1 \bigg  \}.
    \end{aligned}
\end{equation}

\subsection{Our complete objective function}
The final objective of our model is the combination of all the aforementioned loss functions using dedicated Lagrangian coefficients as:
\begin{equation}
 \begin{aligned}
    \mathcal{L}_{GAN}(G_X,_Y,D_X,_Y) &= \mathcal{L}_{adv}(G_x,D_x,X,Y)
     \\ & +  \mathcal{L}_{adv}(G_y,D_y,X,Y)
     \\ & + \lambda_1 \mathcal{L}_{rec}(G_x,G_y,\phi, X,Y)
     \\ & + \lambda_2 \mathcal{L}_{cont}(G_x,G_y,\varphi, X,Y)
     \\ & + \lambda_3 \mathcal{L}^r_{cls}(G_x,G_y,D_x,X,Y)
     \\ & + \lambda_4 \mathcal{L}^f_{feat}(G_x,G_y,D_x,X,Y)
     \\ & + \lambda_5 \mathcal{L}_{cnt}(G_x,G_y,\phi, X,Y)
     \\ & + \lambda_6 \mathcal{L}^f_{id}(G_x,G_y,D_x,X,Y),
 \end{aligned}
\end{equation}
where each $\lambda_i$ scales the corresponding objective to achieve better results. $ G_x$ and $G_y $ are the generators, responsible for synthesising images for their respective domains; $ \{x_i\}_{i=1}^{n}$ $\in \mathcal {X}$ for images and $ \{y_i\}_{i=1}^{n}$ $\in \mathcal {Y} $ for sketches. Discriminators  $ D_x $ and $ D_y $ critic the images in domain $  \mathcal{X}$ and $ \mathcal{Y} $, respectively. $\phi$ represents the overall feature maps we concatenate to the images before computing the reconstruction loss. Our objective function basically aims to solve the min-max game:
 
\begin{equation}
\begin{aligned}
  G_X^*,G_Y^* = arg \min_{(G_x,G_y)}\max_{(D_x,D_y)}  \mathcal{L}_{GAN}(G_x,G_y,D_x,D_y).
    \end{aligned}
\end{equation}


\section{Training}
The models were trained using an Adam optimizer, with $\beta_1 $ = 0.5 and  $\beta_2$  = 0.999. We used a batch sizes ranging between 8 and 16 for all experiments on CelebA \cite{liu2015faceattributes}, WVU Multi-modal \cite{WVU} and CelebA-HQ \cite{CelebAMask-HQ} for RGB images  and from an auxiliary generator trained on sketches from CUHK \cite{li2012human}, IIT-D \cite{Bhatt2010OnMS} and FERET \cite{Phillips1998TheFD}  datasets  and a learning rate of $10^{-5}$ for the first 10 epochs which linearly decayed to 0 over the next 10 epochs.  We trained the entire model on  three NVIDIA Titan X Pascal GPUs.


To train our proposed sketch-to-photo GAN synthesizer, we prioritize our objective to synthesize images from sketches by conditioning each sketch image with a uniquely generated attribute label to form a new channel depth (i.e., 8-channels) as the input channels to the sketch-to-image generator and discriminator (see Figure 3). The input image which now comprises of 8 channels at the input is fed into the discriminator, our hybrid discriminator improves the attribute learning scheme by steering the model towards producing photo-realistic images from the sketches in the dataset.
Our novel quality guided encoder $E_x$ learns the quality features of all the real images at different user specified stages along its encoding network. Its corresponding encoding stage at the generator $G_x$( with shared weights) is linked via an $L_2$ loss to guide image syntheses towards more quality refined images. 

\subsection{Dataset description}
We synthesised sketches using XDOG filter for the CelebA, WVU Multi-modal datasets and from an auxiliary generator trained for sketches from CUHK , IIT-D and FERET datasets. 
The CelebFaces Attributes (CelebA) is a large scale  dataset of  202,599  celebrity  face  images,  each annotated with 40 attributes and 10,177 identities.  We randomly  select  2,000  images  as  the  test  set  and  use  all  the remaining images  as  the  training  data.   
We  generate five domains using the following attributes:  hair color (black, blond, brown), age (young/old). The WVU Multi modal dataset contains 3453 high resolution color frontal images of 1200 subjects. The FERET dataset consists of over 14,126 images for 1199 different persons, with  a variety in face poses, facial expressions, and lighting conditions. The total number of frontal and near frontal face images, whose pose angle lies between -45 and +45, is 5,786 images (3,816 male images and 1,970 female images). IIIT-D sketch dataset contains 238 viewed pairs, 140 semi-forensic pairs, and 190 forensic pairs, while CUHK Face Sketch dataset (contains 311 pairs).

\subsection{Evaluation}
To evaluate the performance of our approach, we compared a set of images against a gallery of mugshots utilizing a synthesised image probe. The gallery comprises of WVU Multi-Modal, CUHK, FERET, IIIT-D and CelebA-HQ datasets. The purpose of this experiment is to assess the verification performance of the proposed method with a relatively large number of subject candidates, Figures (6, 7, 8 and 9) show CMC curves for CelebA, CUHK and IIIT-D datasets, respectively. 

We compare our method with several state  of the art  sketch to image synthesis methods as shown in Figures (7, 8 and 9). To evaluate the performance of the synthesized images, we implemented a face verifier called DeepFace \cite{Taigman2014DeepFaceCT}, pre-trained on a VGG based network \cite{Simonyan14c}. A similar protocol implemented in \cite{8347106} was used.



\subsection{Qualitative assessment}

Ablation analysis on our loss functions were implemented to test the independent efficacy on the model in general. By multiplying the adversarial loss by a degrading values of $\lambda_i$, we observed a considerable effect on the entire performance of the model, as  $\lambda_1$ arrived at zero, the generative potential of the network dwindled accordingly. We also evaluated the cycle loss in only one direction; GAN and forward cycle loss  or GAN and backward cycle loss and found out that it often incurs training instability and causes mode collapse. Due to the fact that the attributes are concatenated at the generator's end, it was inferred that increasing our weights affects the discriminative power of the network especially for the attribute generation for target images. Table 1 and 2 show the computed Fréchet Inception Distance (FID), Feature Similarity index FSIM, null-space linear discriminant analysis (NLDA),  Inception Score (IS) and the Structural Similarity (SSIM) index. 
Table 3 gives a score assessment of the ablation study for the GAN performance metric used.
\begin{figure}
\centering
\begin{minipage}[c]{\textwidth}
\centering
    \includegraphics[width=3.0in]{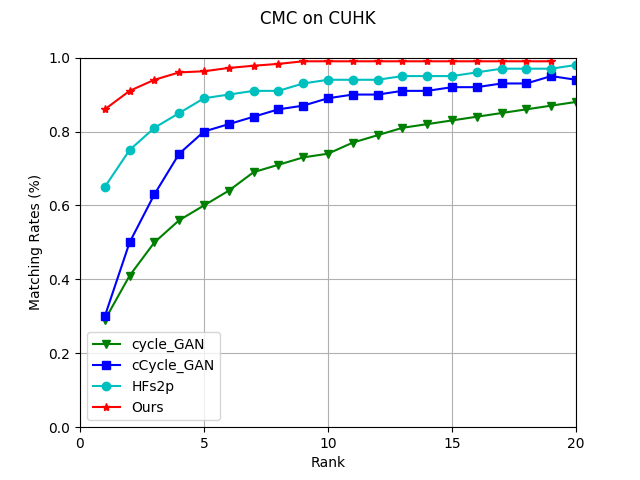}
    \caption{CMC curves of our framework against cycleGAN, cCycleGAN, HFs2P algorithm for the CUHK dataset.}
    \label{fig:sample_figure}
\end{minipage}
\end{figure}

\begin{figure}
\centering
\begin{minipage}[c]{\textwidth}
\centering
    \includegraphics[width=3.0in]{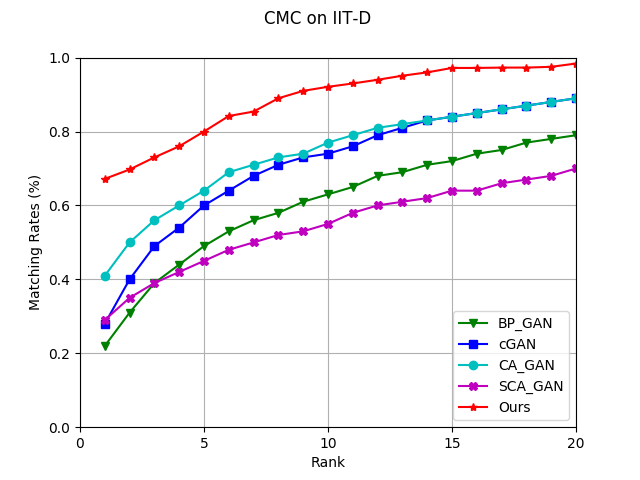}
    \caption{CMC curves of our framework against bpGAN, caGAN, scAGAN and cGAN algorithm for the CelebA dataset.}
    \label{fig:sample_figure}
\end{minipage}
\end{figure}

\begin{figure}
\centering
\begin{minipage}[c]{\textwidth}
\centering
    \includegraphics[width=3.0in]{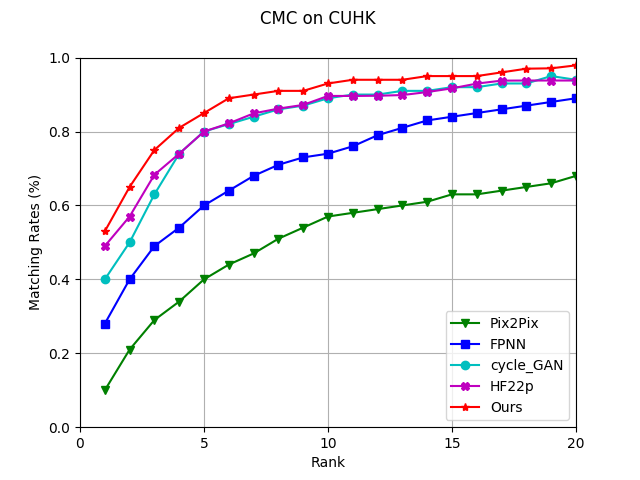}
    \caption{CMC curves of our framework against Pix2Pix, cycleGAN, HFs2P and FPNN algorithm for the IIIT-D dataset.}
    \label{fig:sample_figure}
\end{minipage}
\end{figure}

\vspace{0.5cm}
\begin{figure}
\centering
\begin{minipage}[c]{\textwidth}
\centering
    \includegraphics[width=6.5cm, height=5.1cm]{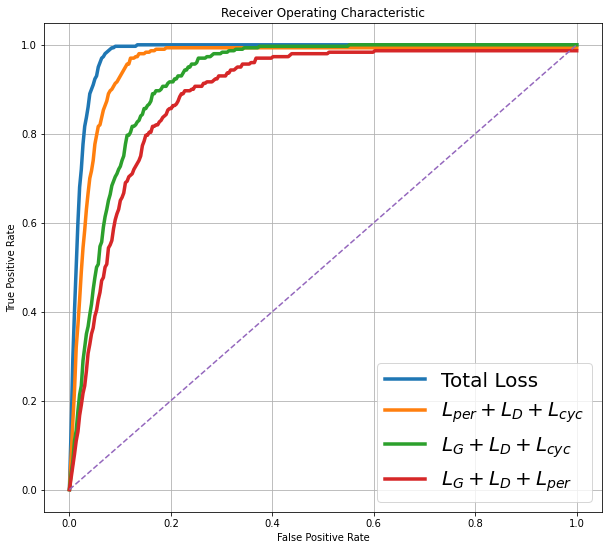}
    \caption{We show the ROC curves showing the importance of different loss functions for ablation study.}
    \label{fig:sample_figure}
\end{minipage}
\end{figure}

\begin{table}
\centering
\begin{adjustbox}{width=7.2cm}
\small

\begin{tabular}{ccccc}

\hline
\multirow{2}{*}{ Metric} & \multicolumn{3}{c}{Models}  \\
\cline{2-4} & FID $\downarrow$ & IS  $\uparrow$& SSIM  $\uparrow$  \\ 
\hline
\hline
Pix2Pix\cite{8100115}     & 72.18   & 1.35 $\pm$ .03  &  0.67 $\pm$ .01 \\  
HFs2p\cite{Chao2019HighFidelityFS}      & 60.21   & 1.48 $\pm$ .03  &  0.79 $\pm$ .01 \\ 
C-GAN\cite{CycleGAN2017}   & 67.13   & 2.79 $\pm$ .01  &  0.74 $\pm$ .04 \\ 
cCycle-GAN\cite{8347106}  & 45.39   & 3.40 $\pm$ .07  &  0.83 $\pm$ .07 \\
Ours        & 37.46   & 3.63 $\pm$ .01  &  0.89 $\pm$ .06 \\
\end{tabular}
\end{adjustbox}
\caption{
A quantitative comparison of the GAN-metric performance for  Pix2Pix, cCycle-GAN, C-GAN, HFs2P and ours. Our proposed approach shows an improvement overall.
}
\vspace{.5cm}
\end{table}

\begin{table}
\centering
\begin{adjustbox}{width=6cm}
\small
\begin{tabular}{ccccc} 
\hline
\multirow{2}{*}{ Metric} & \multicolumn{3}{c}{Models}  \\
\cline{2-4} & FID $\downarrow$ & FSIM  $\uparrow$ & NLDA $\uparrow$ \\ 
\hline
\hline 
BP-GAN\cite{Wang2017BackPA}    & 86.1   &  69.13  &  93.1 \\ 
C-GAN\cite{CycleGAN2017} & 43.2   &  71.1   &  95.5 \\
CA-GAN\cite{Yu2017TowardsRF}    & 36.1   &  71.3   &  95.8 \\
SCA-GAN\cite{Yu2017TowardsRF}   & 34.2   &  71.6   &  95.7 \\
Ours      & 34.1   &  72.8   &  97.0 \\
\end{tabular}
\end{adjustbox}
\caption{
A quantitative comparison of the GAN-metric performance for  BP-GAN, CA-GAN,  SCA-GAN,  C-GAN  and ours. Our proposed approach shows an improvement overall.
}
\vspace{.3cm}
\end{table}
\begin{table}
\centering
\begin{adjustbox}{width=8cm, height=1.3cm}
\small
\begin{tabular}{ccccccccc} 
\hline 
\multirow{2}{*}{Loss} & \multicolumn{3}{c}{Model Resolution} & \multicolumn{3}{c}{Metric} \\
\cline{2-7} & {256 x 256} & {128 x 128} & {64 x 64}  & {FID $\downarrow$} & {IS$\uparrow$ } & {SSIM$\uparrow$ } \\ 
\hline
\hline
$  \mathcal{L}_{G }$ &\cmark & \xmark & \cmark & 37.46 & 5.48 & 0.738 \\ 

$  \mathcal{L}_{D} $ &\xmark & \cmark & \cmark & 31.86 & 2.06 & 0.791 \\ 

$  \mathcal{L}_{cyc} $ &\cmark & \xmark & \cmark & 33.73 & 5.91 & 0.808 \\ 

$  \mathcal{L}_{per} $ &\cmark & \xmark & \cmark & 35.34 & 6.32 & 0.896   \\ 
\end{tabular}
\end{adjustbox}
\caption{
\label{tab:table-name}
A description of the ablation study conducted on the sketch-photo-synthesizer network. The various key components that make up the framework were altered to identify their respective impact on the GAN metric performance (i.e, FID, SSIM and IS).
}
\vspace{0.3cm}
\end{table}

\section{Conclusion}

In this paper, we proposed a novel sketch-to-image translation model using a hybrid discriminator and a multi-stage generator. Our model shows that the perceptual appeal of sketches can be achieved with the network reconfiguration of the generator and discriminator processes. The breaking down of the functionality of the network into smaller subsets helped to  improve the training process and hence led to better results under short periods. Our verification results indeed confirm that sketch-to-image translation problems would find lots of applications in industry.

{\small
\bibliographystyle{ieee_fullname}
\bibliography{egpaper_final}
}

\end{document}